# diffIRM: A Diffusion-Augmented Invariant Risk Minimization Framework for Spatiotemporal Prediction over Graphs

Zhaobin Mo
Department of Civil Engineering and Engineering Mechanics, Columbia University
zm2302@columbia.edu

Haotian Xiang
Department of Electrical Engineering, Columbia University
hx2341@columbia.edu

Xuan Di*
Department of Civil Engineering and Engineering Mechanics, Columbia University
Data Science Institute, Columbia University
sharon.di@columbia.edu

Spatiotemporal prediction over graphs (STPG) is challenging, because real-world data suffers from the Out-of-Distribution (OOD) generalization problem, where test data follow different distributions from training ones. To address this issue, Invariant Risk Minimization (IRM) has emerged as a promising approach for learning invariant representations across different environments. However, IRM and its variants are originally designed for Euclidean data like images, and may not generalize well to graph-structure data such as spatiotemporal graphs due to spatial correlations in graphs. To overcome the challenge posed by graph-structure data, the existing graph OOD methods adhere to the principles of *invariance existence* (i.e., there exist invariant features that consistently relate to the label across various environments), or *environment diversity* (i.e., diversifying training environments increases the likelihood that test environments align with training ones). However, there is little research that combines both principles in the STPG problem. A combination of the two is crucial for efficiently distinguishing between invariant features and spurious ones. In this study, we fill in this research gap and propose a diffusion-augmented invariant risk minimization (diffIRM) framework that combines these two principles for the STPG problem. Our diffIRM contains two processes: i) data augmentation and ii) invariant learning. In the data augmentation process, a causal mask generator identifies causal features and a graph-based diffusion model acts as an environment augmentor to generate augmented spatiotemporal graph data. In the invariant learning process, an invariance penalty is designed using the augmented data, and then serves as a regularizer for training the spatiotemporal prediction model. We provide theoretical evidence supporting diffIRM's ability to identify invariant features. The effectiveness of diffIRM is further demonstrated through experiments on both numerical and real-world data. The numerical data is generated from a known structural causal model (SCM), and our proposed diffIRM successfully identifies the true invariant features. The real-world experiment uses three human mobility datasets, i.e. SafeGraph, PeMS04, and PeMS08. Our proposed diffIRM outperforms baselines. Furthermore, our model demonstrates interpretability by discerning invariant features while making predictions.

---

*Corresponding author, Tel.: +1-212-853-0435.

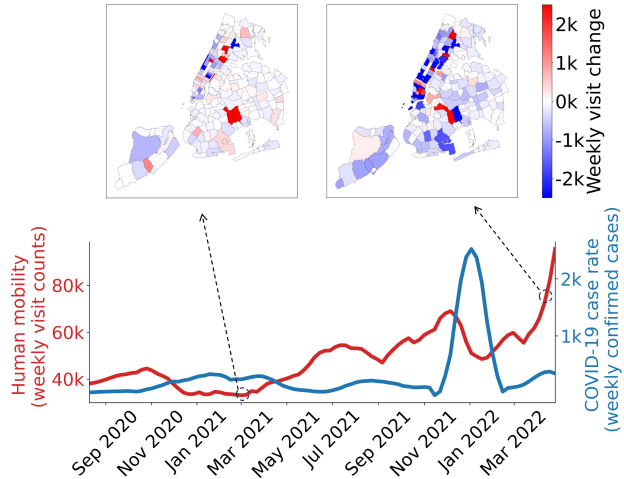

**Figure 1** An illustrative example of the distribution shift for ST graph data. A distribution shift was observed around January of 2022, potentially due to the emergence of the Omicron variant[1]. In other words, the distribution of the weekly visit change manifests different patterns.

## 1. Motivation

Spatiotemporal (ST) prediction over graphs (STPG) aims to unravel intricate patterns and dependencies inherent in spatially distributed entities evolving over time. Its applications span diverse fields, including transportation (Sun 2016), epidemiology (Wang et al. 2022), social networks (Min et al. 2021), among others. In particular, STPG has been extensively studied and applied to a variety of transportation applications, including traffic flow and speed prediction (Feng et al. 2023, Guo et al. 2020b,a, Zhang et al. 2019), ride-hailing demand prediction (Ke et al. 2021a, Feng et al. 2021, Ke et al. 2021b, Tang et al. 2021), next location prediction (Hong et al. 2023), parking occupancy prediction (Yang et al. 2019), and incident prediction (Tran et al. 2023). Understanding and predicting these spatiotemporal dynamics can facilitate informed decision-making (Soppert et al. 2022, Li et al. 2022d), optimal resource allocation (Wang et al. 2023), and enhanced risk mitigation strategies (Li et al. 2021, Bao et al. 2019).

The majority of existing STPG studies, however, are based on the in-distribution hypothesis, i.e., training and test data are drawn from the same distributions. This hypothesis may not hold for real-world data, especially when the data is time-varying. This issue is known as the out-of-distribution (OOD) generalization issue. For example, Fig. 1 illustrates the change in human mobility patterns after the outbreak of COVID-19. The x-axis is the date, and the left and right y-axes indicate the human mobility (i.e. weekly number of visits at the zipcode-level) and the COVID-19 case rates (i.e. weekly confirmed cases), respectively. The red and blue curves represent the evolution of human mobility and the COVID-19 case rate from 2020 August to 2022 April in New York City (NYC). The heatmaps on the top represent the changes in weekly visits, i.e.

---

[1] Source: https://en.wikipedia.org/wiki/Timeline_of_the_COVID-19_pandemic_in_New_York_City#cite_note-139

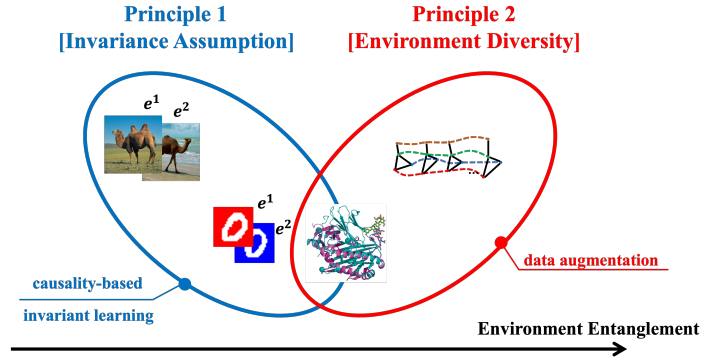

**Figure 2** Illustration of how two principles are related to (1) three graph OOD methods (invariant learning, causality-based, and data augmentation) and (2) two data formats, including the Euclidean (such as images) and graph data (such as protein structure and spatiotemporal graph data). The x-axis represents the increase of environment entanglement from Euclidean data to graph data. Principle 1, encompassing invariant learning and causality-based methods, applies to both Euclidean and graph data. Principle 2, which involves data augmentation techniques, is predominantly utilized for graph data, owing to the increased environmental entanglement. (The hand-writing digit images are from the CMNIST dataset (Arjovsky et al. 2019), and others are from web sources.)

the visit increment from the previous week. Non-stationarity is observed, and we can see both the human mobility and COVID-19 case rate data exhibit apparently different spatiotemporal trends before and after January 2022, when Omicron variant was dominant. In this example, the OOD issue is exemplified by the pattern changes in human mobility and COVID-19 case rates before and after the outbreak of COVID-19. The cause of this OOD issue owes to the change of *environments*, which are the latent factors governing the pattern of the observation, after the surge of COVID-19 case rates.

The OOD problem over graphs is less explored compared to over Euclidean data, such as images. Unlike Euclidean data, where OOD scenarios are often visually discernible through changes in background or color, OOD issues in graph data are inherently more complex due to environment entanglement. This complexity arises from the interconnected nature of graph structures. For instance, in Fig. 2, the environment of images can be easily identified by their visual backgrounds. In contrast, the environment of a graph is more intertwined, affected by the relationships and interactions within the graph's structure itself. Faced with the nuanced challenge of OOD over graphs, current methodologies predominantly adhere to the following two principles.

**Principle 1.** *(Invariance Assumption) Invariant features or representations exist, maintaining a consistent relationship with the label across various environments.*

**Principle 2.** *(Environment Diversity) Diversifying training environments increases the likelihood that the test environment aligns with the training environment.*

To address Principle 1, *causality-based* methods leverage causal inference techniques to identify invariant factors across training and test environments. Also in accordance with Principle 1, *invariant learning* methods (also referred to as robust training) aim to refine the model training procedure and train the model in a manner that minimizes the variability of the training loss across diverse training environments. Unlike causality-based methods, the invariant learning method does not explicitly pinpoint invariant representation, and instead it devises stable training algorithms that leverage the characteristics of invariant features. To realize Principle 2, *data-augmentation* methods aim to generate more training data from diverse environments, thereby encompassing a broader spectrum of environments. In the next section, we will detail each method and the methodological gap in the graph OOD prediction problem.

## 2. Related Work

We first clarify some important concepts that will be discussed throughout the literature review, followed with a toy example to better understand these concepts and the OOD problem.

**Definition 2.1.** *(Environment) An environment, denoted as e, is defined as the latent factor that governs the generation process of features and the label.*

**Definition 2.2.** *(Causal and Environment Features) Denote features and the label as $X$ and $Y$, respectively. Suppose $X$ contains two distinct features, $X_{cau}$ and $X_{env}$, such that $X_{cau} \cup X_{env} = X$. If the condition $Pr(Y \mid X_{cau}) = Pr(Y \mid X)$ holds true, then we categorize $X_{cau}$ as causal features (also known as "invariant" or "rational" features) and $X_{env}$ as environment features (also known as "spurious" features).*

Take Fig. 2 as an example. For the camel pictures, environment $e^1$ is associated with pictures of a camel in the desert and environment $e^2$ is associated with a camel on the beach. In this example, the invariant feature is the camel pixel and the environment feature is the background scenery. Similarly, for hand-writing digital pictures, the invariant feature is digital and the environment feature is the background color. A classification model trained on environment $e^1$ may not generalize effectively if the test data is generated from environment $e^2$. This is because the model may learn the spurious relationship associated with the image background.

**Definition 2.3.** *(Out-of-Distribution) Out-of-Distribution (OOD) is defined as the change in the joint distribution of features and the label between the training and test data, which is caused by the change of environments in the training and test data.*

We use a motivating example to illustrate the OOD problem. Consider the structural causal model (SCM):

$$X_1 \leftarrow \text{Gaussian}(0, \sigma^2),$$
$$Y \leftarrow X_1 + \text{Gaussian}(0, \sigma^2),$$
$$X_2 \leftarrow Y + \text{Gaussian}(0, 1).$$

To formulate an OOD generalization problem, assume that the training data is generated from environments $\mathcal{E}_{\text{train}} = \{\text{replace } \sigma^2 \text{ by } 4, 7\}$, and the test data is generated from environments $\mathcal{E}_{\text{test}} = \{\text{replace } \sigma^2 \text{ by } 8, 9\}$. In this example, $X_1$ is the causal feature and $X_2$ is the environment feature.

Let $D_{\text{train}} = \{x_1^{(i)}, x_2^{(i)}, y^{(i)}, \sigma^{(i)}\}_{i=1}^{N_{\text{train}}}$ and $D_{\text{test}} = \{x_1^{(j)}, x_2^{(j)}, y^{(j)}, \sigma^{(j)}\}_{j=1}^{N_{\text{test}}}$ represent the training and test data, respectively, where $N_{\text{train}}$ and $N_{\text{test}}$ denote the sizes of each dataset. For simplicity, we assume that environments in all datasets are uniformly distributed. For example, $Pr(\sigma = 4) = Pr(\sigma = 7) = 0.5$ for the training data. In this simplified example, we utilize a linear regressor $y = f_\theta(x_1, x_2) = \theta_1 x_1 + \theta_2 x_2$ trained on the training data $D_{\text{train}}$ and tested on the test data $D_{\text{test}}$. Note that the regressor is unaware of the underlying environment $\sigma$ during neither training nor testing.

The solutions of different methods are presented in Tab. 1. The ominous method supposes we know $x_1$ is the causal feature and forces the model solely to learn $x_1$. For ERM, we use least squares and the solution is solved analytically. The discrepancy between the solutions of ERM and Ominous is that inevitably learn the spurious relationship between $x_2$ and $y$. Random augmentation adds random perturbation to the feature matrix, a technique utilized in Kong et al. (2022). Employing diffIRM, the learned regressor is $y = 0.93x_1 + 0.10x_2$, which closely approximates the ground-truth invariant solution $y = x_1$.

Table 1   Solutions for the Motivating Example.

| Method | Solution | Data Augmented? | Causality Preserved? |
|---|---|---|---|
| Ominous | $y = 1.00x_1$ | - | - |
| ERM | $y = 0.11x_1 + 0.89x_2$ | ✗ | ✗ |
| Random Augmentation | $y = 0.34x_1 + 0.53x_2$ | ✓ | ✗ |
| diffIRM (ours) | $y = 0.93x_1 + 0.10x_2$ | ✓ | ✓ |

## 2.1. General OOD Methods for Euclidean Data

Traditional statistical approaches, like least squares, train the prediction model by minimizing the empirical risk on the training data, a process known as Empirical Risk Minimization (ERM) (Vapnik 1991). To address the OOD issue of ERM, the pioneering study of Invariant Risk Minimization (IRM) (Arjovsky et al. 2019) introduces the concept of environments and identifies that OOD arises when both training and test data encompass a mix of distinct environments. IRM penalizes the gradient of the empirical risk with regard to the classifier parameters, assuming that changing the classifier parameters will not impact the model optimality if this model only learns invariant features and discard environment ones.

Since the inception of IRM, many methods (Chang et al. 2020, Zhang et al. 2021, Xu and Jaakkola 2021, Ahuja et al. 2020, Ahmed et al. 2020, Creager et al. 2021, Liu et al. 2021, Krueger et al. 2021) have been introduced to address the OOD issue for Euclidean data, as summarized in Tab. 2. These methods are divided into three categories, based on whether the environments are known, unknown, or inferred. Except for IRM that assumes unknown environments, all other methods either assume the environments are known or can be inferred. The environment-known methods, assuming that the environments data belong to are known, define an invariance penalty based on the performance discrepancy across environments. For example, Risk

Extrapolation (REx) (Krueger et al. 2021) penalizes the variance of the empirical risks among all known environments. This penalization is based on the assumption that the optimal model should have equivalent performance over all environments, thus having a zero empirical risk variance. The environment-inferred methods, in contrast, learn to infer the environment partitions, and define an invariance penalty based on the inferred environments (Ahmed et al. 2020, Creager et al. 2021, Liu et al. 2021, Lin et al. 2022).

Table 2    General OOD methods for Euclidean data.

| Model | Task | Environment Known or Inferred? |
|---|---|---|
| IRM (Arjovsky et al. 2019) | image classification | unknown |
| IRM-Game (Ahuja et al. 2020) | image classification | known |
| TRM (Xu and Jaakkola 2021) | image classification | known |
| REx (Krueger et al. 2021) | image classification | known |
| Transfer (Zhang et al. 2021) | image classification | known |
| InvRat (Chang et al. 2020) | text review classification | known |
| PGI (Ahmed et al. 2020) | image classification | inferred |
| EIIL (Creager et al. 2021) | image classification | inferred |
| HRM (Liu et al. 2021) | price prediction | inferred |
| ZIN (Lin et al. 2022) | image classification | inferred using auxiliary information |

The aforementioned studies, relying on the assumption of knowing or inferring an environment, may not hold for graph-structured data due to the complex spatial correlations in graphs. Taking Fig. 2 for example, invariant and environment features of the camel and digital images can be straightforwardly separated as foreground and background pixels. However, graph data, such as protein and spatiotemporal graphs, often exhibit a high degree of *environment entanglement*, which is characterized by the difficulty of distinguishing between invariant and environment features, making the environments of the graph data expensive or even impossible to obtain. Furthermore, unlike Euclidean data such as images, the definition of the environment on a graph is unclear, which makes the entanglement of environments more challenging to infer and disentangle.

### 2.2. OOD Methods for Graph Data

There is a growing body of literature aiming to address the challenge of OOD over graphs from environment entanglement, primarily categorized into three groups: causality-based approaches, data augmentation techniques, and invariant learning methods, which are summarized in Tab. 3.

Data augmentation approaches (Feng et al. 2020, Liu et al. 2022b, Kong et al. 2022, You et al. 2020, Yu et al. 2022, Liu et al. 2022a, Sui et al. 2022, Li et al. 2022c) follows Principle 2 and are designed to enrich training environments, thereby reducing model overfitting in specific scenarios. These methods generate augmented data by imposing node/edge masks or permutations on the original graph data. For example, some studies (Feng et al. 2020, You et al. 2020, Sui et al. 2022, Li et al. 2022c) employ a binary mask to exclude certain elements of node features or the adjacency matrix, thereby augment node feature values or edge connectivity, respectively. Others introduce trainable perturbations (Kong et al. 2022, You et al. 2020) or generate new subgraphs (Liu et al. 2022a,b, Yu et al. 2022) by training a data augmetor, typically with a multilayer perceptron (MLP). Depending on whether adjacency matrices are augmented, these methods can be further divided into node feature augmentation (Feng et al. 2020, Liu et al. 2022b, Kong et al. 2022) and graph topology augmentation (Li et al. 2022a). In this paper, as our focus is on citywide traffic prediction with a relatively stable graph topology, we concentrate on augmenting node features on a graph while keeping its topology unchanged. While data augmentation is widely used due to its ease of implementation, such a method lacks a theoretical foundation for determining which part of the data should be augmented. Thus, data augmentation may potentially affect invariant features that are supposed to be constant across environments. Also, due to the lack of prior knowledge about the characteristics of invariant and environment features, data augmentation may require a significant amount of time to find the optimal augmentation that solely alters the environment features, making this category of methods less efficient compared to the causality-based and invariant learning methods to be introduced later.

Causality-based methods (Zhou et al. 2022, Fan et al. 2022, Chen et al. 2022, Zhang et al. 2022, Wu et al. 2022b,c, Zhao et al. 2022, Xia et al. 2023) adhere to Principle. 1 by assuming that data is generated from an underlying SCM where the label is only determined by invariant features (i.e., the label's causal parents) and not by environment features. These methods first train a classifier to identify invariant features and then solely rely on the identified invariant features for prediction. While most methods utilize a similar architecture for the classifier, such as MLP or GNN, these methods leverage different properties of invariant features to train a classifier. For instance, some studies (Zhou et al. 2022, Zhang et al. 2022, Wu et al. 2022b,c, Zhao et al. 2022, Xia et al. 2023), based on the idea that the invariant features are causal parents of the label in the SCM, block spurious correlation from environment variables using causal inference techniques, such as front-door or back-door adjustments and counterfactuals. Other studies, instead of using rigorous causal inference theory, rely on prior knowledge. For example, Chen et al. (2022) leverages the property that invariant features have the maximum mutual information concerning the label. Fan et al. (2022) assumes that environment features are easier to learn than invariant ones, and uses this assumption to distinguish between the invariant and environment features. However, relying on an SCM requires discovering a causal diagram beforehand and could be non-trivial for real-world applications.

Invariant learning methods also adhere to Principle 1 by assuming the existence of invariant features. Unlike the aforementioned causality-based methods, invariant learning methods do not assume data generation from an SCM. Instead, invariant learning extends the general OOD methods mentioned in Tab. 2 to graph-structure data. While following the same invariant principle, existing studies differ in how to design a regularization term for the training process. For example, Sadeghi et al. (2021) incorporates robust learning methods into graph neural networks (GNN), regularizing the training process through a worst-case loss.

Wu et al. (2022a) designs a regularization term to minimize the loss variance across environments. Some studies (Li et al. 2022b, Miao et al. 2022) design such regularization terms based on maximizing the mutual information between invariant features and the label. Additionally, some research (Zhu et al. 2021, Buffelli et al. 2022) suggests using a metric, central moment discrepancy, as a regularization term to quantify OOD. However, most methods adopt general invariant learning strategies established for Euclidean data, not graph-structured data.

Table 3　Methods for Graph OOD.

| Model | Task | Categories | | | Principles | |
|---|---|---|---|---|---|---|
| | | Cau. | IL | DA | Inv. Asmp. | Env. Div. |
| GRAND (Feng et al. 2020) | node/edge classification | ✗ | ✗ | ✓ | ✗ | ✓ |
| LA-GNN (Liu et al. 2022b) | node/edge classification | ✗ | ✗ | ✓ | ✗ | ✓ |
| FLAG (Kong et al. 2022) | node/edge classification | ✗ | ✗ | ✓ | ✗ | ✓ |
| GraphCL (You et al. 2020) | node/edge/graph classification | ✗ | ✗ | ✓ | ✗ | ✓ |
| DPS (Yu et al. 2022) | node/graph classification | ✗ | ✗ | ✓ | ✓ | ✓ |
| GREA (Liu et al. 2022a) | node/edge/graph classification | ✓ | ✗ | ✓ | ✓ | ✓ |
| AdvCA (Sui et al. 2022) | graph classification | ✓ | ✗ | ✓ | ✓ | ✓ |
| RGCL (Li et al. 2022c) | graph classification | ✓ | ✗ | ✓ | ✓ | ✗ |
| gMPNN (Zhou et al. 2022) | link prediction | ✓ | ✗ | ✗ | ✓ | ✗ |
| DisC (Fan et al. 2022) | graph classification | ✓ | ✗ | ✗ | ✓ | ✗ |
| CIGA (Chen et al. 2022) | graph classification | ✓ | ✗ | ✗ | ✓ | ✗ |
| DIDA (Zhang et al. 2022) | link prediction | ✓ | ✗ | ✗ | ✓ | ✗ |
| DIR (Wu et al. 2022b) | graph classification | ✓ | ✗ | ✓ | ✓ | ✗ |
| DSE (Wu et al. 2022c) | graph classification | ✓ | ✗ | ✓ | ✓ | ✗ |
| CFLP (Zhao et al. 2022) | link prediction | ✓ | ✗ | ✓ | ✓ | ✓ |
| CaST (Xia et al. 2023) | ST prediction | ✓ | ✗ | ✗ | ✓ | ✗ |
| GNN-DRO (Sadeghi et al. 2021) | node prediction | ✗ | ✓ | ✗ | ✓ | ✗ |
| GIL (Li et al. 2022b) | graph classification | ✗ | ✓ | ✗ | ✓ | ✗ |
| SR-GNN (Zhu et al. 2021) | node classification | ✗ | ✓ | ✗ | ✓ | ✗ |
| SSReg (Buffelli et al. 2022) | node classification | ✗ | ✓ | ✗ | ✓ | ✗ |
| GSAT (Miao et al. 2022) | graph classification | ✗ | ✓ | ✓ | ✓ | ✓ |
| EERM (Wu et al. 2022a) | graph classification | ✗ | ✓ | ✓ | ✓ | ✓ |
| StableGL (Shengyu et al. 2023) | node classification | ✗ | ✓ | ✓ | ✓ | ✓ |
| diffIRM (ours) | ST Prediction | ✓ | ✓ | ✓ | ✓ | ✓ |

"Cau.": causality-based; "IL": invariant learning; "DA": data augmentation; "Inv. Asmp.": invariance assumption; "Env. Div.": environment diversity.

Integration of Principles 1 and 2 is crucial for enabling the model to effectively discern invariant representations, while simultaneously distinguishing between diverse environments. While the integration of these two principles is explored in some studies (Yu et al. 2022, Liu et al. 2022a, Sui et al. 2022, Zhao et al. 2022, Miao et al. 2022, Wu et al. 2022a, Shengyu et al. 2023), these studies mainly focus on classification tasks, and no study combines these two principles for the task of STPG.

Addressing these research gaps, we propose a novel framework, **Diff**usion-augmented **I**nvariant **R**isk **M**inimization (diffIRM), which strategically integrates the merits of causality-based, invariant learning, and data augmentation methods. Unlike the general OOD methods outlined in Tab. 2, diffIRM adopts a unique approach by utilizing data augmentation to generate environments, rather than relying on inference or pre-assumed knowledge of these environments. To maintain the integrity of the causality structure in the augmented environments, we identify causal features through a min-max game. In this game, a causal mask generator aims to identify the causal features, while an environment augmenter works to diversify the environmental features. The environments generated from this game are instrumental to invariant learning, specifically by aiding in the formulation of an invariant penalty.

Our contributions are summarized as follows:

1. Our proposed diffIRM is the first to combine the above-mentioned two principles in STPG. Additionally, diffIRM combines the merits of three graph OOD methods, i.e. causality-based, invariant learning, and data augmentation.
2. We provide theoretical evidence that diffIRM can identify causal features.
3. To evaluate diffIRM's effectiveness, we conduct extensive experiments, including a numerical motivating example and real-world datasets, specifically Safegraph, PeMS 04 and PeMS08 datasets.

The rest of this paper is organized as follows. Sec. 3 introduces the preliminaries and the problem statement. Sec. 4 fleshes out the framework of our proposed diffIRM, along with its proof and training algorithm. Sec. 5 details the experiments and presents the results. Sec. 6 concludes our work and project future directions.

## 3. Preliminaries and Problem Statement

This section will introduce the preliminaries and problem statement. Tab. 4 introduces the major notations that will be used in this paper.

### 3.1. Preliminaries

**3.1.1. IRM.** IRM (Arjovsky et al. 2019) decomposes a prediction model $f(\cdot)$ into a feature extractor $h$ and a predictor $g$, represented as $f = g \circ h$. IRM addresses the OOD issue by solving a bi-level optimization problem:

$$\min_{g,h} \sum_{e \in \mathcal{E}_{tr}} \mathcal{L}^e(g \circ h),$$
$$\text{subject to} \quad g \in \arg\min_{g^e} \mathcal{L}^e(g^e \circ h), \forall e \in \mathcal{E}_{tr}, \tag{3.1}$$

where $\mathcal{E}_{tr}$ represents the training environment and $\mathcal{L}^e(f) = \mathbb{E}_{X^e, Y^e}[\ell(f(X^e), Y^e)]$ calculates the empirical risk of environment $e$. Eq. 3.1 aims to learn an invariant feature extractor $h(\cdot)$ such that the predictor $g(\cdot)$ is simultaneously optimal across training environments.

Table 4  Summary of Notation Used in the Paper

| Notation | Description |
| --- | --- |
| $X$ | features |
| $Y$ | label |
| $D$ | dataset consisting of features and label |
| $\bar{X}$ | latent features in the diffusion process |
| $\hat{X}$ | generated features from the diffusion model |
| $\tilde{X}$ | augmented features |
| $f$ | ST prediction model |
| $T$ | causal mask generator |
| $G$ | environment augmentor |
| $\theta$ | parameters of the ST prediction model $f$ |
| $\phi$ | parameters of the causal mask generator $T$ |
| $\psi$ | parameters of the environment augmentor |
| $\mathcal{L}$ | empirical risk |
| $\ell$ | loss function |
| $r$ | invariance penalty |

**3.1.2. GNN.** In an undirected graph $\mathcal{G} = (V, E, A)$, $V = \{v_i\}_{i=1}^{N}$ represents the set of nodes, where $N = |V|$ is the number of nodes. The set $E$ represents edges, and $A \in \mathbb{R}^{N \times N}$ denotes the adjacency matrix, which indicates the node connectivity. The entry $A_{ij}$ in the matrix indicates the presence of an edge, where $A_{ij} = 1$ if edge $(v_i, v_j) \in E$, otherwise $A_{ij} = 0$. Node features are represented as $X$.

GNNs generalize the neural network technique to graph data, allowing information propagation between nodes and their neighbours. For example, in graph convolutional networks (GCN), the propagation rule is defined as $H^{(l+1)} = \sigma(\hat{A} H^{(l)} W^{(l)})$, where $\hat{A}$ is the symmetric normalized adjacency matrix, $\sigma()$ represents the ReLU activation function, $W^{(l)}$ is the weight matrix of the $l^{\text{th}}$ layer, and $H^{(l)}$ is the latent node representation of the $l^{\text{th}}$ layer with $H^{(0)} = X$. GNNs can also incorporate attention mechanisms, such as in attention-based spatiotemporal GCN (ASTGCN), in which an adaptive adjacency matrix $A_{adp}$ is used alongside the static adjacency matrix $A$ to dynamically adjust the impact weights between nodes. The adaptive adjacency matrix is defined as $A_{adp} = W_1 \cdot \sigma\left((XW_2) W_3 (W_4 X)^T + b\right)$, where $W_{1:4}$ and $b$ are learnable parameters.

Built upon GNNs, ST-GNNs further extend to temporal dimensions. In most ST-GNN architectures, GNNs are utilized for capturing spatial relations, while traditional temporal models, like the Long Short-Term Memory (LSTM) (Xue et al. 2022) and gated Temporal Convolutional Network (gated TCN) (Yu et al. 2017, Wu et al. 2019, Lan et al. 2022), are used to capture temporal relations. For example, the spatiotemporal GCN (STGCN) uses a GCN and a gated TCN to capture spatial and temporal relations, respectively. More recently, the attention mechanism has been integrated into ST GNNs for temporal modeling, such as in ASTGCN.

## 3.2. Problem Statement

Denote $X_t^i \in \mathbb{R}^F$ as the feature of node $i$ at time step $t$, and $X_t = [X_t^1, X_t^2, \cdots, X_t^N] \in \mathbb{R}^{N \times F}$ as the values of all nodal features at time $t$. We further denote $X_{(t-\tau+1):t} = [X_{t-\tau+1}, \cdots, X_t] \in \mathbb{R}^{N \times \tau \times F}$ as the historical nodal features from the previous $\tau$ time steps, and $Y_{(t+1):(t+\tau')} = [Y_{t+1}, \cdots, Y_{t+\tau'}] \in \mathbb{R}^{N \times \tau' \times 1}$ as the future node features of length $\tau'$. Assume that there is an underlying mapping $f(\cdot)$ from the historical nodal features to future ones: $X_{(t-\tau+1):t} \xrightarrow{f(\cdot)} Y_{(t+1):(t+\tau')}$. For conciseness, we refer to $X_{(t-\tau+1):t}$ as $X$ and $Y_{(t+1):(t+\tau')}$ as $Y$ in the rest of this paper. The problem of STPG considering the OOD issue can be defined as:

**Problem 1.** *(STPG with OOD) In the problem of spatiotemporal prediction over a graph considering OOD issues, the goal is to learn a function $f_\theta(\cdot)$ that is simultaneously optimal over all environments:*

$$\min_\theta \sup_{e \in \mathcal{E}} \mathcal{L}^e(f_\theta),$$

where $\mathcal{L}^e(f_\theta) = \mathbb{E}_{(X^e, Y^e) \sim D^e}[\ell_\theta(f_\theta(X^e), Y^e)]$ is the empirical risk of environment $e$ and $\ell : \mathbb{R}^{N \times \tau \times F} \times \mathbb{R}^{N \times \tau \times F} \to \mathbb{R}_{\geq 0}$ is the error measurement such as the mean squared error (MSE). The subscript of $\ell$ indicates the parameter that is related to calculating $\ell$.

To learn a function $f_\theta$ that is optimally effective across various environments, it is essential to identify features of which relationship with the label is invariant, such as the camel and digital pixels in the example from Fig. 2.

## 4. Methodology

Our proposed diffIRM model consists of two primary components (as illustrated in Fig. 3): *data augmentation* and *diffusion-augmented IRM*. During the data augmentation process, the original data is fed into a causal mask generator $T_\phi$ and an environment augmentor $G_\psi$ to generate a total of $K$ sets of augmented data. The role of the causal mask generator $T_\phi$ is to differentiate between causal and environmental features, while the environment augmentor $G_\psi$ aims to increase the diversity of environmental features. During the diffusion-augmented IRM process, as the previous data augmentation effectively diversifies the environment features, the generated data can be perceived as originating from distinct environments. Leveraging this explicit environment separation, we define an invariance penalty to compel the STPG model $f_\theta$ to learn the mapping between the causal features and the label. Our model can be formulated as a min-max game:

$$\min_\theta \left\{ \min_\phi \left[ \max_\psi \underbrace{\mathbb{E}_{(\tilde{X}, Y) \sim \tilde{D}} \ell_{\theta, \phi, \psi}\left(f_\theta(\tilde{X}), Y\right)}_{\text{augmentation loss}} \right] + \underbrace{r(\theta)}_{\text{invariance penalty}} \right\}, \quad (4.1)$$

where $\tilde{X}$ is a augmented feature and $\tilde{D}$ is the dataset consisting of the augmented feature $\tilde{X}$ and unaugmented label $Y$. The data augmentation module consists of an environment augmentor $G_\psi$ and a causal mask generator $T_\phi$. $r(\theta)$ is the invariance penalty.

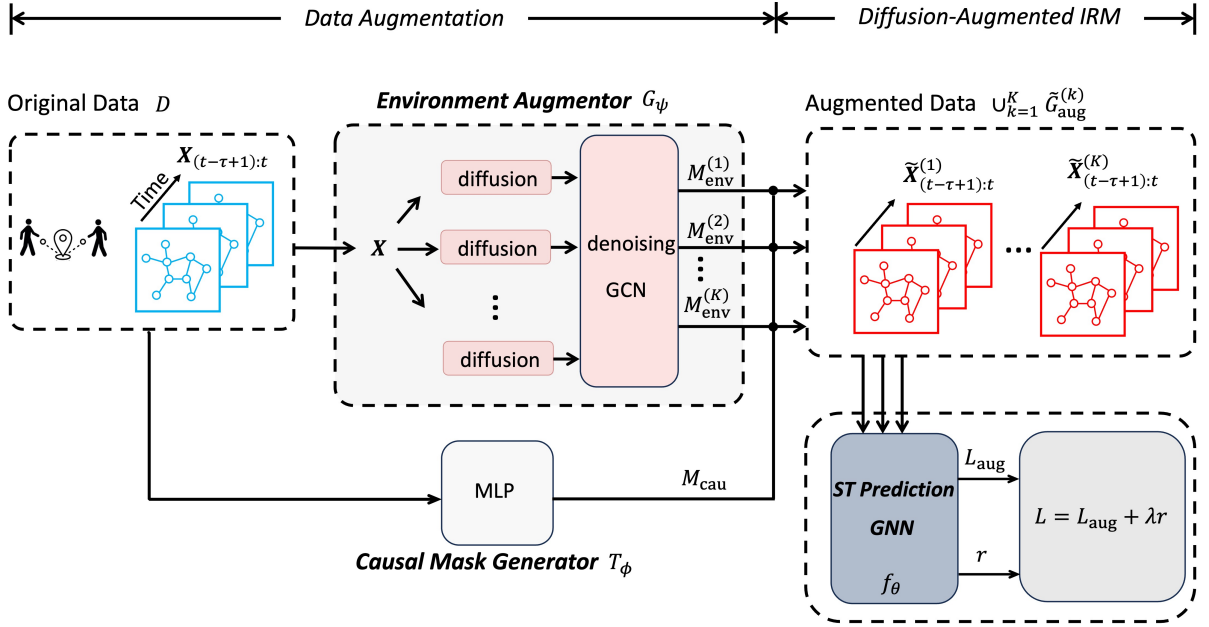

**Figure 3**  Overview of the proposed diffIRM.

A detailed description of data augmentation and invariant learning are provided in Sec. 4.1 and Sec. 4.2, respectively. In Sec. 4.3, we provide theoretical evidence that diffIRM can identify causal features. In Sec. 4.4, we revisit the motivating example mentioned in Sec. 2 and demonstrate that the augmented data retain the same causal and environment features as the original one. The training algorithm will be introduced in Sec. 4.5.

### 4.1. Data Augmentation

Given a fixed $\theta$, the min-max game $\min_\phi \left[ \max_\psi \mathbb{E}_{(\tilde{X},Y)\sim \tilde{D}} \ell_{\theta,\phi,\psi} \left( f_\theta(\tilde{X}), Y \right) \right]$ consists two components: a data augmentor $G_\psi$ and a causal mask generator $T_\phi$. On the other, the $G_\psi$ aims to augment environment features so that the objective function $\mathbb{E}_{(\tilde{X},Y)\sim \tilde{D}} \ell_{\theta,\phi,\psi} \left( f_\theta(\tilde{X}), Y \right)$ is maximized, resulting in a diversified data environment. On the other, the goal $T_\phi$ is to identify causal features. In the following two subsections, we will provide detailed explanations of how the augmented feature $\tilde{X}$ is generated using $G_\psi$ and $T_\phi$.

**4.1.1. Environment Augmentation using Diffusion.** We employ a diffusion model $G_\psi : \mathbb{R}^{N\times\tau\times F} \times [0,1]^{N\times N} \rightarrow \mathbb{R}^{N\times\tau\times F}$ as an environment augmentor. Denote $\hat{X} = G_\psi(X, A)$ as the augmented node feature. In contrast to the original diffusion model, we use a GCN instead of a U-Net as the neural backbone, as GCN can capture spatial correlation using the pre-calculated adjacency matrix $A$. Our diffusion model is based on the DDPM (Ho et al. 2020), which has demonstrated remarkable performance in image generation tasks. Notably, DDPM has also been applied to tasks related to graph data such as protein structure generation (Trippe et al. 2022). DDPM comprises a forward diffusion process and a reverse denoising process. In the diffusion process, the original feature is corrupted with Gaussian noise over $L_{\text{diff}}$ steps. Each diffusion step follows the conditional distribution

$$q(\bar{X}^{(l)}|\bar{X}^{(l-1)}) = \mathcal{N}(\bar{X}^{(l)}; \sqrt{1-\alpha^{(l)}}\bar{X}^{(l-1)}, \alpha^{(l)}I), \tag{4.2}$$

where $\bar{X}^{(l)}$ represents the corrupted feature after $l$-step diffusion with $l \in \{1, ..., L_{\text{diff}}\}$ and $\bar{X}^0 = X$, $\alpha^{(l)}$ controls the strength of the $l$-step diffusion, and $I$ is an identity matrix. With proper design of $\alpha^{(l)}$ and sufficient diffusion steps, the final corrupted feature $\bar{X}^{(L_{\text{diff}})}$ approximate a standard Gaussian distribution. In the reverse denoising process, the diffusion model aims to predict the noise based on the corrupted feature so that the noise can be removed, which follows the following conditional distribution parameterized by $\psi$

$$p_\psi(\bar{X}^{(l-1)}; \mu_\psi(\bar{X}^{(l)}, l, A), \Sigma_\psi(\bar{X}^{(l)}, l, A)), \tag{4.3}$$

where both $\mu_\psi(\cdot)$ and $\Sigma_\psi(\cdot)$ are GCNs. Compared to the original GCN introduced in the preliminaries Sec. 3.1.2, both $\mu_\psi(\cdot)$ and $\Sigma_\psi(\cdot)$ are fed in with $l$ as an additional input. The diffusion step $l$ is input to compute the position embedding, which uses the Transformer sinusoidal position embedding method (Vaswani et al. 2017). We denote the final reconstructed feature as $\hat{X} := \bar{X}^{(0)}$. For the simplification of the notation, we represent the augmentation of the diffusion model in the form of matrix multiplication: $\hat{X} = X \odot M_{\text{env}}$, where $M_{\text{env}} \in \mathbb{R}^{N \times \tau \times F}$ is a mask parameterized by $\psi$.

**4.1.2. Causal Mask Generation.** While the diffusion model may alter causal features, we address this challenge by introducing a multi-layer perception (MLP) $T_\phi : \mathbb{R}^{N \times \tau \times F} \to [0, 1]^{N \times \tau \times F}$ to identify causal features. Denote $M_{\text{cau}} = T_\phi(X)$ as a mask that indicates whether a particular element $X_{i,j,k}$ corresponds to a causal feature ($[M_{\text{cau}}]_{i,j,k} = 1$) or not ($[M_{\text{cau}}]_{i,j,k} = 0$). We employ a soft mask for stable training. Combining both environment augmentation and causal mask generation, we define the entire data augmentation as the mask matrix $M_{\psi,\phi} = M_{\text{env}} \odot M_{\text{cau}}$. Thus, the final augmented feature $\tilde{X}$ can be expressed as

$$\begin{aligned}\tilde{X} &:= X \odot M_{\psi,\phi} = X \odot M_{\text{cau}} + \hat{X} \odot (1 - M_{\text{cau}}) \\ &= X \odot M_{\text{cau}} + X \odot M_{\text{env}} \odot (1 - M_{\text{cau}}).\end{aligned} \tag{4.4}$$

It is worth noting that $M_{\text{cau}}$ is generated deterministically from an MLP, while $M_{\text{env}}$ is stochastically determined as a sample from the diffusion model's outputs. This design aligns with the assumption that causal features should remain stable (otherwise they are not invariant across environments), whereas environmental features exhibit randomness across different environments. Specifically, we use $M_{\psi,\phi}^{(k)} = M_{\text{env}}^{(k)} \odot M_{\text{cau}}$ to represent a single augmentation sampled from the diffusion model. We denote the augmented dataset associated with this augmentation as $\tilde{D}^{(k)} = \{(X^i \odot M_{\psi,\phi}^{(k)}, Y^i)\}_{i=1}^N$. With the unified data augmentation matrix $M_{\psi,\phi}$, we generate a total of $K$ augmented environments from the original dataset $D$, resulting in a augmented dataset denoted as $\tilde{D} = \cup_{k=1}^{K} \tilde{D}^{(k)}$. Using this augmented data, the *augmentation loss* in Eq. 4.1 can be detailed as:

$$\mathcal{L}_{\text{aug}}(\theta, \psi, \phi) = \mathbb{E}_{(\tilde{X},Y)\sim \tilde{D}} \ell_{\theta,\phi,\psi}(\tilde{X}, Y) \approx \mathbb{E}_{(X,Y)\sim D} \sum_{k=1}^{K} \ell_{\theta,\phi,\psi}(f_\theta(X \odot M^{(k)}_{\psi,\phi}), Y)/K. \quad (4.5)$$

The detailed training algorithm for updating parameters will be in Sec. 4.5.

### 4.2. Diffusion-Augmented IRM

Now we discuss the construction of the invariance penalty in Eq. 4.1 using the augmented dataset $\tilde{D}$. The core idea behind the invariant penalty is that the environment augmentor $G_\psi$ is trained to maximize the empirical risk by altering $X_{\text{env}}$, thereby making the augmented feature $X \odot M_{\psi,\phi}$ as diverse as possible in terms of environments. Thus, we can define the invariant penalty as the error discrepancy between $f_\theta$ and an environment-specific predictor $f_{\theta_k}(\cdot)$, where parameter $\theta_k$ is optimized using data solely from $\tilde{D}^{(k)}$. The mathematical form of the invariant penalty is

$$r(\theta) = \mathbb{E}_{(X,Y)\sim D} \sum_{k=1}^{K} [\ell_{\theta,\phi,\psi}(f_\theta(X \odot M^{(k)}_{\psi,\phi}), Y) - \ell_{\theta_k,\phi,\psi}(f_{\theta_k}(X \odot M^{(k)}_{\psi,\phi}), Y)]/K. \quad (4.6)$$

Note that $\ell_{\theta_k,\phi,\psi}(f_{\theta_k}(X \odot M^{(k)}_{\psi,\phi}), Y)$ is the lower bound of $\sum_{k=1}^{K} \ell_{\theta,\phi,\psi}(f_\theta(X \odot M^{(k)}_{\psi,\phi}), Y)/K$, because the former can been seen as overfitting the predictor $\theta_k$ on data $\tilde{D}^{(k)}$. Thus, $r(\theta) = 0$ if and only if the predictor $f_\theta$ is spontaneously optimal in all augmented environments, in which case $f_\theta$ is the invariant predictor by the definition of invariant preditor in Sec. 3.1.1, where the invariant feature extractor $h(\cdot)$ become the invariant predictor $f(\cdot)$ with $g(\cdot) = 1$. The final loss function for updating $\theta$ is:

$$\mathcal{L}(\theta) = \mathcal{L}_{\text{aug}}(\theta, \psi, \phi) + \lambda r(\theta), \quad (4.7)$$

where $\lambda$ is a hyperparameter that controls the weight of the invariant penalty. The training algorithm for updating $\theta$ will be elaborated in Sec. 4.5.

### 4.3. Causal Feature Identifiability

Now we will analyze diffIRM from a theoretical perspective. We aim to prove that the loss function in Eq. 4.7 exhibits the property of *causal feature identifiability* (CFI). CFI is characterized by a loss function, by minimizing which causal features can be distinguished from the environment ones.

**Definition 4.1.** *(Causal Feature Identifiability (CFI)) Let $\mathcal{L}_{cau}$ represent the loss value of Eq. 4.7 when $f_\theta$ successfully learns causal features $X_{cau}$ alone. Let $\mathcal{L}'$ denote the loss value of a feature $X' \subset X$ with $X' \neq X_{cau}$. The loss function can be said to exhibit CFI if $\mathcal{L}_{cau} < \mathcal{L}'$.*

CFI ensures that training $f_\theta$ with Eq. 4.7 enables $f_\theta$ to identify causal features. This is crucial because we ultimately use the predictor $f_\theta$ to forecast future graph features. Thus, it is imperative to demonstrate that our designed loss function empowers the predictor $f_\theta$ to discern causal features, enhancing its ability to generalize to unseen test data. Then, we introduce the condition that needs to be met to achieve CFI.

**Condition 1.** *(CFI Condition) When the min-max game (Eq. 4.1) between the mask generator $T_\phi$ and the environment augmentor $G_\psi$ reaches its equilibrium, the causal mask generator is able to identify causal features $X_{cau}$.*

Condition 1 requires that the trained causal mask generator generates the correct causal feature mask, i.e. $M_{cau} = M_{cau}^* = [\mathbf{1}^{d_v}, \mathbf{0}^{d_s}]$, where $d_v$ and $d_s$ denote the dimension of $X_{cau}$ and $X_{env}$, respectively. Our proof will focus on proving that Condition 1 is both the sufficient and necessary condition for diffIRM to achieve CFI. In the appendix, we will discuss how Condition 1 can be met.

**4.3.1. Condition 1 → CFI.** First, we would like to show that Condition 1 is a sufficient condition for CFI. For simplicity, we omit the subscript of $M_{\psi,\phi}$ and use $M$ to represent the data augmentation. Denote the optimal value of $\mathbb{E}_{(X,Y)\sim D}[\sum_{k=1}^K \ell_{\theta,\phi,\psi}(f_\theta(X \odot M_{\psi,\phi}^{(k)}), Y)/K]$ as $\mathcal{L}^*(Y|X, M^{(1:K)})$, which represents the lower bound of the loss function when employing a neural network to predict $Y$ using the augmented data $\tilde{D}$. Similarly, denote the optimal value of $\mathbb{E}_{(X,Y)\sim D}[\ell_{\theta_k,\phi,\psi}(f_{\theta_k}(X \odot M_{\psi,\phi}^{(k)}), Y)]$ as $\mathcal{L}(Y|X, M^{(k)})$, which corresponds to the optimal loss associated with learning from a specific augmented dataset $\tilde{D}^{(k)}$. We use subscripts to denote the sub-feature and its corresponding augmentation, e.g., $X_1 \subset X$ and $\tilde{X}_1 = X_1 \odot M_1$.

**Assumption 1.** *(Loss lower bound) For any constant $\epsilon > 0$, there exist a $\theta$ such that $\mathbb{E}[\ell(f_\theta(X), Y)] \leq \mathcal{L}^*(Y|X) + \epsilon$; Additionally, for all constant $\epsilon > 0$, $\mathbb{E}[\ell(f_\theta(X), Y)] \geq \mathcal{L}^*(Y|X) - \epsilon$.*

$\mathcal{L}^*$ indicates the minimum of the risk. The first part of Assumption 1 is the universal approximation theorem of neural networks, i.e. the expressiveness of neural networks is enough so that they can approximate any functions. The second part pertains to the property of the lower bound.

**Assumption 2.** *For any distinct features $X_1$ and $X_2$, $\mathcal{L}^*(Y|X_1) - \mathcal{L}^*(Y|X_1, X_2) \geq \gamma$ with fixed $\gamma \geq 0$.*

Assumption 2 indicates that any feature can provide some useful information for predicting $Y$ that cannot be explained by other features. $\gamma$ can be 0 when $X_2$ provide nothing useful for prediction $Y$, e.g. $X_2$ is white noise.

**Assumption 3.** *If a feature $X_s$ violates the invariance constraint, adding another feature $X$ would not make the penalty vanish. That is, there exists a constant $\sigma > 0$ so that $\mathcal{L}^*(Y|X_s, X, M^{(1:K)}) - \mathcal{L}^*(Y|X_s, X, M^{(k)}) \geq \delta(\mathcal{L}^*(Y|X_s, M) - \mathcal{L}^*(Y|X_s, M^{(k)}))$*

Assumption 3 aims to ensure a sufficient positive penalty given the existence of a environment feature.

**Assumption 4.** *For any feature $X_1$ and a corresponding augmentation $M_1$, continually augmenting the same feature with an additional augmentation $M_1'$ will increase the optimal loss. That is, $\mathcal{L}^*(Y|X_1, M_1, M_1') - \mathcal{L}^*(Y|X_1, M_1) \geq C$, with fixed $C > 0$.*

Assumption 4 holds because, unlike the inclusion of additional features as in Assumption 2, the augmentation is based on the existing feature and will not provide new information about providing $Y$. Including additional augmentation equivalently diversifies environments such that fitting those environments with a single function $f_\theta$ leads to more discrepancy and thereby a larger optimal loss.

With the aforementioned Assumptions 1-4 and Condition 1, we present the following theorem:

**Theorem 1.** *If $\lambda > \frac{\mathcal{L}^*(Y)+2\epsilon}{\delta C - 4K\epsilon}$ and $\epsilon < \min\{\frac{\delta C}{4K}, \frac{\gamma}{2+4\lambda K}\}$, we conclude that $\mathcal{L}^*_{cau} < \mathcal{L}^*_{env+}$. Thus, training the prediction model $f_\theta$ with Eq. 4.7 leads to the identification of invariant features.*

Theorem 1 indicates that Condition 1 is sufficient for identifying invariant features using Eq. 4.7. A proof is provided in the appendix.

### 4.3.2. $\overline{\text{Condition 1}} \to \overline{\text{CFI}}$.
Now we prove that Condition 1 is also a necessary condition for diffIRM to achieve CFI. In other words, if Condition 1 does not hold, identifiability does not hold either.

**Proposition.** *If Condition 1 is violated, the training of the neural network $f$ with the proposed loss function will exclude some causal features. That is, if there exists a causal mask $M^{(k)}$ that augments the original causal features, then there exists some feature $X' \subset X$ with $X' \neq X_{cau}$ and $\mathcal{L}_{cau} > \mathcal{L}'$, where $\mathcal{L}'$ is the loss value associated with $X'$.*

Proposition 4.3.2 indices that if Condition 1 is violated, training the prediction model $f$ with Eq. 4.7 will exclude some causal features. The proof is provided in the appendix.

### 4.4. Revisiting the Motivating Example

We will provide a more detailed explanation of Condition 1, particularly demonstrating that our augmented data $(\tilde{X}, Y)$ maintain the same causal relationships as in the original data. Recall from the motivation example that $X_1$ and $X_2$ are identified as causal and environmental features, respectively, according to Definition 2.2. Therefore, we aim to show that $\tilde{X}_1$ remains causal features, with $Pr(Y|\tilde{X}_1) = Pr(Y|\tilde{X}_1, \tilde{X}_2)$. Furthermore, as discussed in the original IRM paper (Arjovsky et al. 2019), a causal feature should also ensure that $\mathbb{E}[Y|X_{cau}]$ remains invariant across different environments, which will also be demonstrated.

1. $\boldsymbol{Pr(Y|\tilde{X}_1) = Pr(Y|\tilde{X}_1, \tilde{X}_2)}$. Fig. 4(a) shows the distribution of $Pr(Y|\tilde{X}_1)$, while Fig. 4(b-c) depict $Pr(Y|\tilde{X}_1, \tilde{X}_2)$ for three different values of $X_2$. These figures demonstrate that variations in $X_2$ do not affect the distribution of $Pr(Y|\tilde{X}_1, \tilde{X}_2)$, which remains equivalent to $Pr(Y|\tilde{X}_1)$. For enhanced visualization, Fig. 5 presents the conditional expectation $\mathbb{E}[Y|X_1]$ along with its standard deviation. The dashed line represents the expectation according to $Pr(Y|\tilde{X}_1)$ upon marginalizing $X_2$, and the solid lines indicate the expectation values for three distinct $X_2$ scenarios. The observed discrepancy in the lines for $\tilde{X}_2 = -5$ and $\tilde{X}_2 = 5$ can be attributed to numerical errors arising from the use of the Monte Carlo method to estimate the expectation. This is due to there being fewer samples for $\tilde{X}_2 = -5$ and

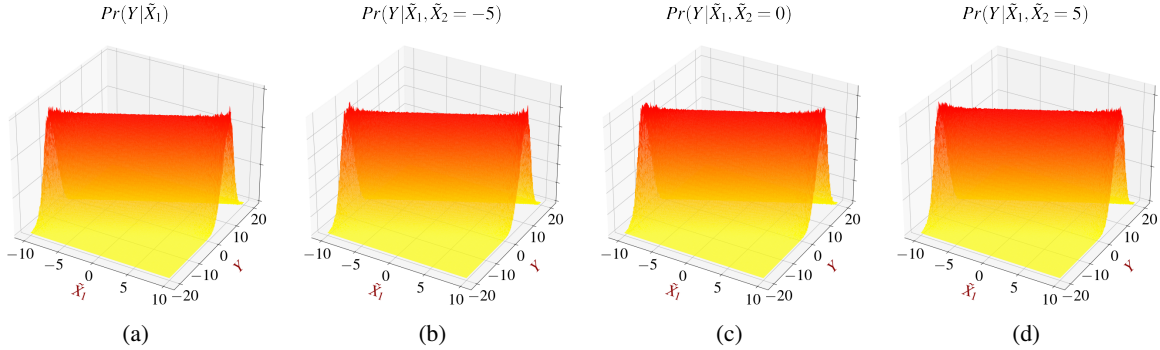

**Figure 4** Validation of $Pr\left(Y \mid \tilde{X}_1\right) = Pr(Y \mid \tilde{X}_1, \tilde{X}_2)$. **(a)** is the conditional distribution $P(Y|\tilde{X}_1)$. **(b-d)** represent the conditional distributions $P(Y|\tilde{X}_1, \tilde{X}_2)$ with $\tilde{X}_2$ equaling to -5, 0, 5, respectively.

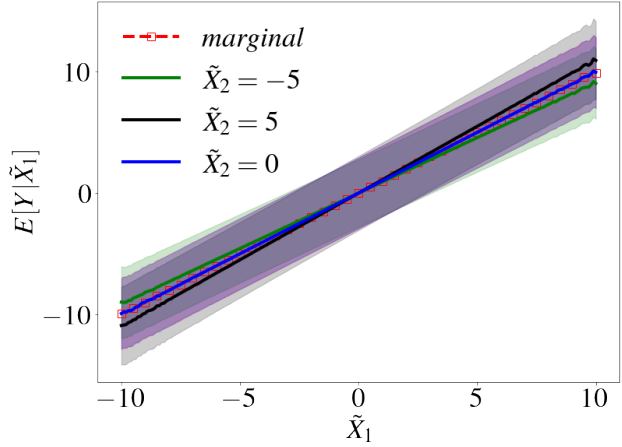

**Figure 5** Validation of $P\left(Y \mid \tilde{X}_1\right) = P(Y \mid \tilde{X}_1, \tilde{X}_2)$

$\tilde{X}_2 = 5$ compared to the scenario where $\tilde{X}_2 = 0$. It is evident that the value of $X_2$ has no effect on the conditional expectation either.

2. **Invariant $\mathbb{E}[Y|X_1]$**. In the motivating example, we use the standard deviation $\sigma$ to represent the environment. Fig. 6 demonstrates the values of $\mathbb{E}[Y|X_1]$ for different $\sigma$. From this illustration, it is clear that the changes in the environment, as signified by different values of $\sigma$, do not impact the value of this expectation.

### 4.5. Training Algorithm

The training algorithm for updating $\psi$, $\phi$, and $\theta$ is depicted in Algorithm 1.

## 5. Experiment

In this section, we will evaluate the performance of our proposed diffIRM using two real-world spatio-temporal graph datasets. We compare diffIRM with several baselines, and also conduct ablation studies to understand the influence of different components within our framework. We also demonstrate that our framework is GNN agnostic and can be applied to different spatiotemporal prediction models

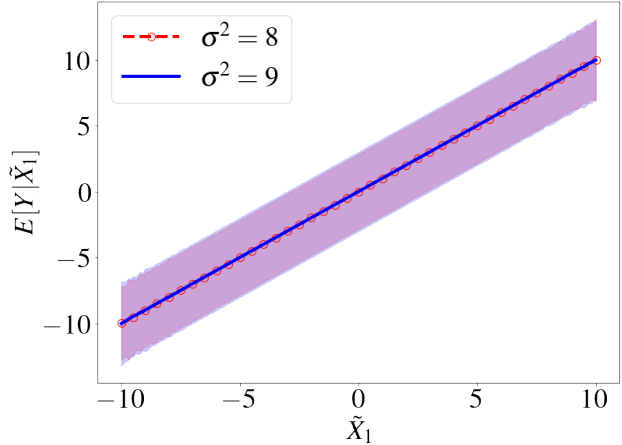

**Figure 6**    Illustration of the invariance of $\tilde{X}_1$.

### 5.1. Experiment Setting

**Dateset**. Three datasets are used in our experiment, i.e. SafeGraph, PeMS04, and PeMS08.

1. *Safegraph* is an open dataset comprising weekly data on the number of visits across 172 zipcode regions in New York, spanning from 08/10/2020 to 04/18/2022. The dataset is aggregated on a weekly basis and has a total of 90 time steps. The dataset also integrates conditional information including weekly COVID-19 confirmed case rates from the Centers for Disease Control and Prevention, and demographic data (regional income and population) sourced from the U.S. Census. Specifically, the node features include visit counts, COVID-19 confirmed cases, regional income, population, weekly precipitation, and points of interest. We divide the dataset into three segments: the first 56 time steps as the training set, the subsequent 16 time steps as the validation set, and the final 16 time steps as the test set. We use the historical 3 time-step data to predict the future 3 time-step one.

2. *PeMS* is an open dataset primarily used for traffic prediction. Traffic data, including flow, density, and occupancy, is collected every 5 minutes using loop detectors. Additionally, we utilize the following aggregated contextual features:
   - Vehicle Miles Traveled (VMT), which is the total vehicle mileage divided by the population, calculated from all loop detectors;
   - Travel Time Index (TTI), indicating the ratio of travel time in peak periods to that in free-flow conditions;
   - daily aggregated counts of road incidents;
   - lane closures.

**Algorithm 1** diffIRM Training Algorithm

**Initialization**:

Initialized parameters for the environment augmentor $\psi$, causal mask generator $\phi$, and STPG model $\theta$; training iterations $Iter$; batch size $m$; learning rate $lr$; historical length $\tau$; prediction interval $\tau'$; invariance penalty weight $\lambda$

**Input**: The processed input-output pairs $\{(X,Y)^{(i)}\}_{i=1}^{|D|}$, where $|D|$ is the dataset size.

1: **for** $iter \in \{0,...,Iter\}$ **do**
2:     Sample batches $B = \{(X,Y)^{(i)}\}_{i=1}^{m}$.
3:     **for** $(X,Y)^{(i)} \in B$ **do**
4:         $M_{\text{cau}} \leftarrow T_\phi(X^{(i)})$
5:         **for** $k \in \{1,...K\}$ **do**
6:             $M_{\text{env}}^{(k)} \leftarrow G_\psi(X^{(i)})$
7:             $M_{\psi,\phi}^{(k)} \leftarrow M_{\text{cau}} + M_{\text{env}}^{(k)} \odot 1 - (M_{\text{cau}})$
8:         **end for**
9:         calculate the augmentation loss $\mathcal{L}_{\text{aug}}(\theta,\psi,\phi)$ using Eq. 4.5
10:        calculate the invariance penalty $r(\theta)$ using Eq. 4.6
11:        calculate the updated values of the parameters of causal mask generator $T_\phi$ by gradient descent:
          $\phi' \leftarrow \phi - \gamma \text{Adam}(\phi, \nabla_\phi \mathcal{L}_{\text{aug}}(\theta,\psi,\phi))$
12:        calculate the updated values of the parameters of environment augmentor $G_\psi$ by gradient descent:
          $\psi' \leftarrow \psi + \gamma \text{Adam}(\psi, \nabla_\psi \mathcal{L}_{\text{aug}}(\theta,\psi,\phi))$
13:        calculate the updated values of the parameters of ST prediction model $f_\theta$ by gradient descent:
          $\theta' \leftarrow \theta - \gamma \text{Adam}(\theta, \nabla_\theta(\mathcal{L}_{\text{aug}}(\theta,\psi,\phi) + \lambda r(\theta)))$
14:        update the parameters:
          $\phi \leftarrow \phi'; \psi \leftarrow \psi'; \theta \leftarrow \theta'$
15:     **end for**
16: **end for**

These aggregated features are computed across all loop detectors daily, meaning each detector shares the same set of contextual features for every time step within a day. In addition to the previously mentioned aggregated features, the dataset includes 13 other ones. However, some features might be repetitive, like the data in the "Mobility Performance Report," primarily composed of layered statistical data regarding traffic speeds. Our experiment uses two segments of the dataset: PeMS 04, encompassing the Bay Area with 307 nodes from 01/01/2018 to 02/28/2018, and PeMS 08, covering the San Bernardino/Riverside area with 170 nodes from 07/01/2016 to 08/31/2016. Each node corresponds to an individual loop detector. For both the PeMS 04 and PeMS 08 datasets, we use a 60/20/20 ratio to split the training, validation, and test data temporally. We use the historical 12 time-step data to predict the future 12 time-step one.

**Baseline**. We compare diffIRM with ERM, IRM, REx, and InvRat. We use the same ASTGCN model for $f_\theta$ for each baseline. Also, we use the same network architecture and hyperparameters for all the baselines, and tune them on the validation set. To benchmark the impact of the contextual feature, we also use an additional autoregressive baseline, $ERM_{AR}$, where only the time-lagged feature is used for training the ASTGCN. More baseline details, including their loss functions, are included in the appendix. Apart from those baselines, we compare different variants of diffIRM to ablate the contribution of each component, they are:

- **AdvAug**. In this variant, both the causal mask $M_{cau}$ and the environment augmentation $M_{env}$ are generated by MLPs, and no invariance penalty is implemented.
- **DiffAug**. DiffAug has the same causal mask generator and environment augmetor as diffIRM, but it does not use an invariance penalty.
- **diffIRM$^-$**. In diffIRM$^-$, the diffusion model is used to generate the augmented data without using the causal mask generator.

To show that diffIRM is GNN agnostic, we also ablate the neural backbone of $f_\theta$ and try other models like LSTM and STCGN, which will be detailed in the coming ablation study part.

**Metrics**. We use mean absolute error (MAE), root mean square error (RMSE), and mean absolute percentage error (MAPE) as evaluation metrics.

Table 5    Evaluation of different models using real-world mobility data

| | Safegraph | | | | PeMS04 | | | | PeMS08 | | | |
|---|---|---|---|---|---|---|---|---|---|---|---|---|
| | average | | final | | average | | final | | average | | final | |
| | MAE | RMSE | MAE | RMSE | MAE | RMSE | MAE | RMSE | MAE | RMSE | MAE | RMSE |
| $ERM_{AR}$ | 892.7 | 1824.4 | 953.5 | 1942.2 | 24.3 | 38.4 | 29.8 | 45.7 | 18.9 | 29.2 | 22.7 | 34.1 |
| ERM | 828.0 | 1509.7 | 876.8 | 1604.8 | 23.9 | 37.4 | 29.0 | 44.2 | 18.4 | 28.4 | 21.8 | 33.0 |
| IRMv1 | 468.7 | 1076.8 | 518.0 | 1208.7 | 22.7 | 35.6 | 26.8 | 41.1 | 18.1 | 28.2 | 22.0 | 33.5 |
| REx | 434.2 | 957.2 | 503.7 | 1129.0 | 23.1 | 36.8 | 28.1 | 43.0 | 18.1 | 28.3 | 22.1 | 33.6 |
| InvRat | 838.4 | 1596.4 | 1130.6 | 2095.3 | 23.7 | 37.7 | 29.1 | 43.9 | 18.5 | 28.4 | 22.0 | 33.6 |
| AdvAug | 232.4 | 453.4 | 278.3 | 593.5 | 22.4 | 35.1 | 26.0 | 39.7 | 17.2 | 26.9 | 20.1 | 32.5 |
| DiffAug | 221.1 | 438.6 | 254.6 | 524.4 | 22.2 | 34.9 | 26.1 | 40.1 | 17.4 | 27.2 | 20.4 | 32.9 |
| diffIRM$^-$ | 122.4 | 292.3 | 132.4 | 308.6 | 21.4 | 33.7 | 24.9 | 38.2 | 16.7 | 25.2 | 19.1 | 29.8 |
| **diffIRM** | **102.5** | **223.4** | **103.6** | **263.5** | **21.2** | **33.3** | **24.5** | **37.8** | **16.1** | **24.8** | **18.8** | **28.3** |

## 5.2. Performance Comparison

**Comparison between our methods and baselines**. Tab. 5 provides a comparative analysis of the performance of diffIRM in relation to baseline models, utilizing the SafeGraph, PeMS04, and PeMS08 datasets.

These results demonstrate that diffIRM achieves superior performance over the baselines across all datasets, particularly when considering different prediction intervals. Fig. 7 depicts the heatmap visualization of the relative errors across different zipcode areas on the map of New York City. The relative error displayed in the heatmap represents the overall relative error for the 3-week-ahead prediction.

**Comparison among our method variants**. We also compare the performance of different variants of diffIRM, such as AdvAug, DiffAug, and diffIRM$^-$. Tab. 5 shows that DiffAug outperforms AdvAug, indicating that the diffusion model is a better way of augmenting the environment features than the mask multiplication. Results also show that diffIRM outperforms diffIRM$^-$, indicating that the causal mask generator is an important component of diffIRM, which can help identify the causal features.

**Impact of degree of OOD**. It is important to note that real-world data is rarely perfectly in-distribution, especially for human mobility spatiotemporal data, which generally encounters different degrees of OOD issues. Our proposed methods demonstrate significant improvements for the SafeGraph data, where the Covid-19 pandemic has altered data patterns in the test set. The improvement is more moderate for PEMS 04 and PEMS 08, as the OOD issue is less severe compared to SafeGraph. However, our method still shows considerable improvement, as real-world data consistently suffers from OOD challenges.

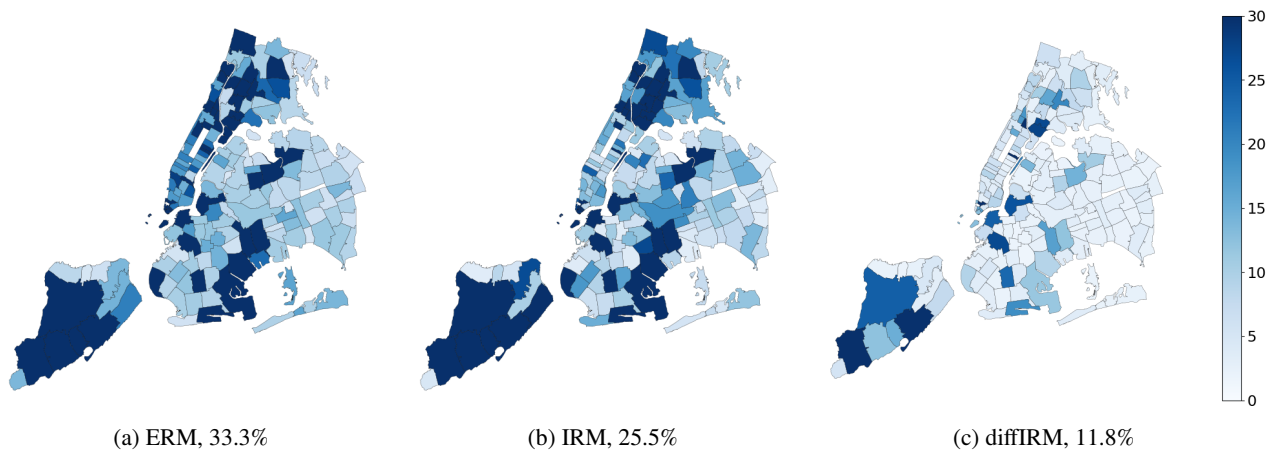

(a) ERM, 33.3%  (b) IRM, 25.5%  (c) diffIRM, 11.8%

**Figure 7** MAPE heatmaps and overall MAPEs of different models for the Safegraph data.

## 5.3. Visualization

Below we visualize the results of the SafeGraph and PeMS 04 datasets.

**Time-series prediction.** Fig. 8 presents the performance of our proposed diffIRM model on the Safegraph dataset for two distinct regions identified by their zip codes: 10003 and 10016. The time series data encompasses weekly visits over 90 weeks, with the last 16 weeks serving as the forecast horizon for our evaluation.

The diffIRM prediction aligns well with the observation for both ZIP codes, showing the model's capability to capture the trends well. Similarly, Fig. 9 illustrate the performance of the diffIRM model applied on PeMS 04 dataset over one day. The chart compares the reconstructed traffic flow values against the ground truth data, as recorded by Sensor 36. The close overlap of the two lines demonstrates the model's high accuracy in reconstructing traffic flow patterns. Notably, diffIRM is able to capture the pattern during better than baselines. ERM differs the most from IRM during transitions in traffic flow patterns, such as at 5:00 (transition from late-night to morning peak) and at 22:00 (transition from evening peak to midnight). This may be because the training of diffIRM makes ASTGCN more aware of the underlying mode changes, allowing it to better capture and adapt to these transitions, whereas ERM struggles to account for such dynamic shifts.

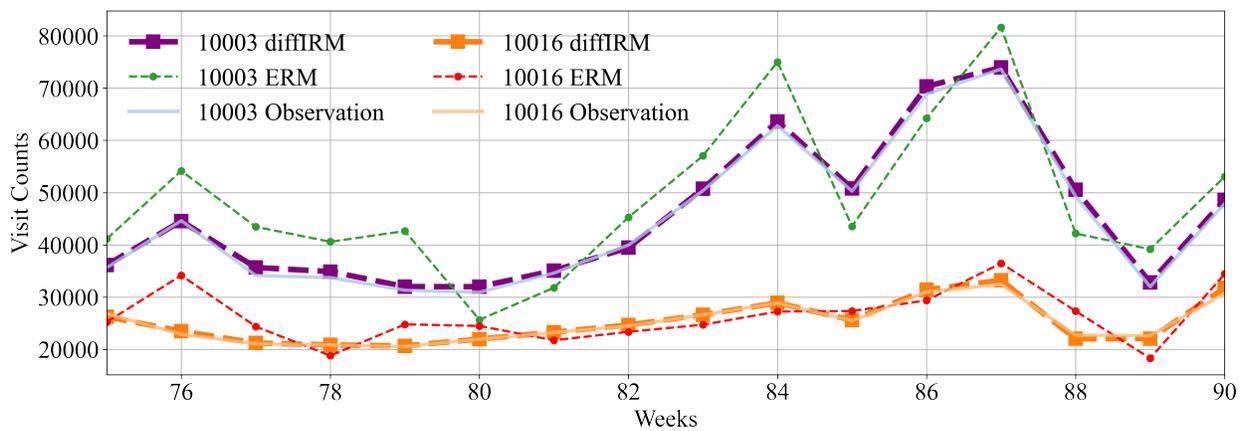

Figure 8   Prediction of visit counts for Safegraph data.

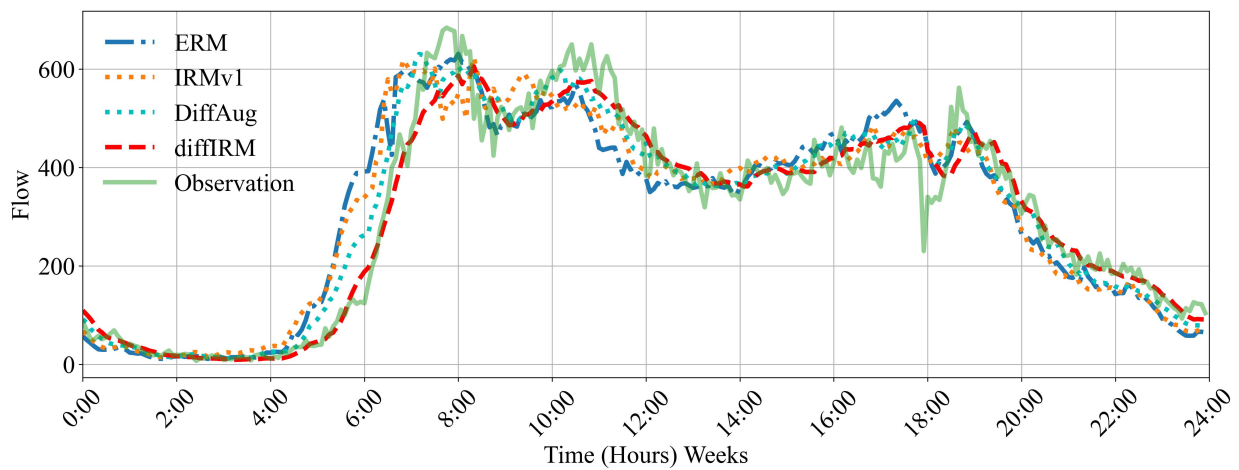

Figure 9   Prediction of traffic flow for Sensor 36 of PeMS 04 data.

**Causal mask**. Fig. 10 visualizes the learned causal masks for the SafeGraph data before and after the outbreak of the epidemic. These matrices are aggregated using one-month data. The x-axis is the time step and the y-axis indicates features. The color represents how certain the causal mask generator $T_\phi$ considers each feature as a causal one at different time steps, with a lighter color indicating higher certainty. Comparing Fig. 10 (a) and (b), we can see that confirmed cases are more likely to be causal features after the outbreak of COVID-19, reflecting that people's mobility pattern is more impacted by the epidemic situation after the confirmed cases surge.

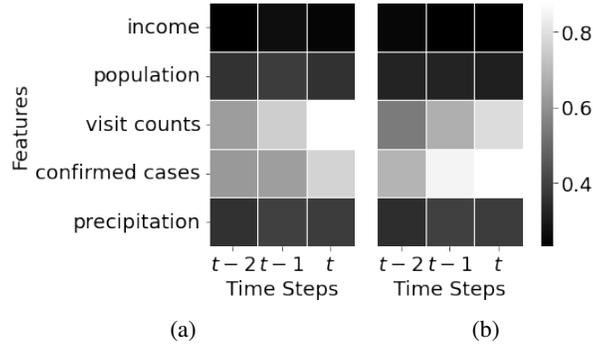

**Figure 10** Causal masks of January 2021 (a) and January 2022 (b) for SafeGraph data.

### 5.4. Ablation Studies

Now we use the SafeGraph data to conduct ablation studies.

**ST prediction models.** Fig. 11 compares the performance between ERM and diffIRM using different STPG models (also known as neural backbones). For the LSTM model, the application of diffIRM reduces the RMSE and MAE by nearly 30%, demonstrating its ability to improve predictions. The STGCN model also shows significant improvement with diffIRM. The ASTGCN model also achieves a significant performance boost from diffIRM, with both RMSE and MAE dropping to less than half of the ERM values. This result shows that diffIRM is agnostic to the STPG model.

**The number of generated environments.** The corresponding ablation study in Tab. 6 shows the impact of the number of environments generated from the diffusion models. Results show that 5 generated environments suffice for enhancing the performance of IRM. Employing additional environments offers only marginal improvements to the model's performance. This suggests that the diffusion model can effectively augment the environment features with a moderate number of environments, and adding more environments may introduce unnecessary noise or redundancy.

**Historical length**. Tab. 7 shows the performance of diffIRM, diffIRM$^-$ and ERM with different historical lengths. This result shows that the advantage of diffIRM over baselines persists under different historical lengths.

**Prediction Steps**. We show the changes in prediction performance as the prediction steps increase in Fig. 12, where Fig. 12(a) and (b) present the changes in RMSE and MAE for different prediction steps, respectively.

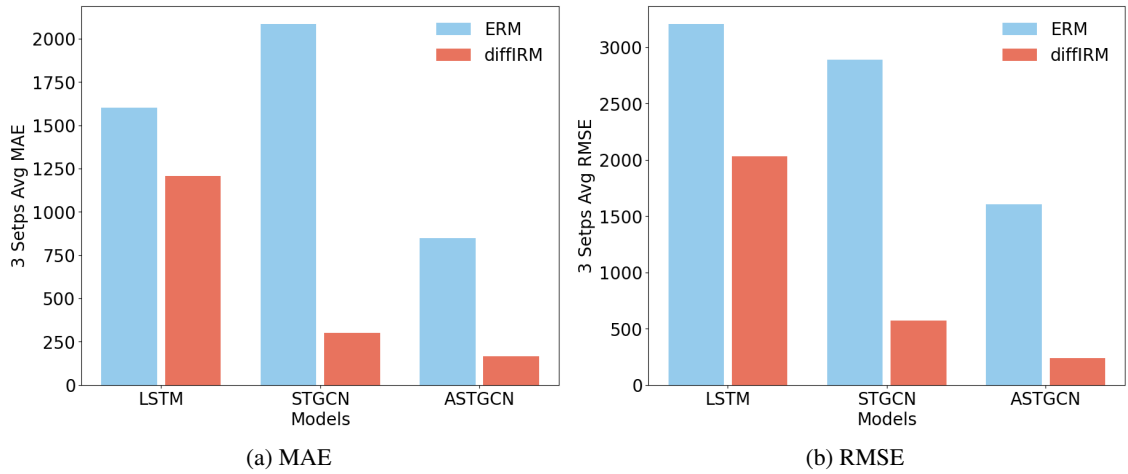

(a) MAE  (b) RMSE

Figure 11  Ablation study of ST prediction GNN categories.

Table 6  Ablation study of different numbers of environments with diffIRM

| Safegraph | 3 envs | 5 envs | 7 envs | 9 envs |
|---|---|---|---|---|
| 3 steps MAE | 169.3 | 132.4 | 144.5 | 160.4 |
| 3 steps RMSE | 843.7 | 308.6 | 587.5 | 843.1 |

These models are trained with the objective of predicting the next 12 time steps using the historical 12 time steps, and the error for each prediction interval is calculated. We can see that diffIRM achieves the best overall performance across all time steps.

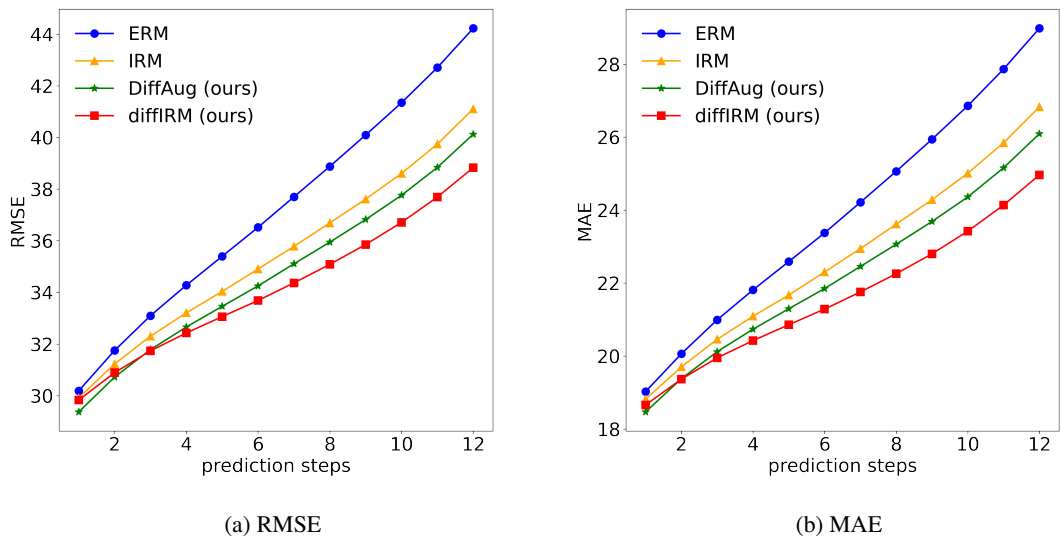

(a) RMSE  (b) MAE

Figure 12  Errors of different prediction steps for PeMS 04 data.

Table 7  Ablation study of historical length $\tau$

|  | $\tau=1$ | | $\tau=2$ | | $\tau=3$ | |
|---|---|---|---|---|---|---|
|  | MAE | RMSE | MAE | RMSE | MAE | RMSE |
| ERM | 934.5 | 1734.5 | 894.3 | 1646.7 | 828.0 | 1509.7 |
| diffIRM$^-$ | 436.3 | 945.3 | 358.2 | 645.6 | 122.4 | 292.3 |
| diffIRM | 463.6 | 973.6 | 258.6 | 486.5 | 102.5 | 223.4 |

## 6. Conlusion

This paper addresses the out-of-distribution (OOD) issue in the problem of spatiotemporal prediction over graphs (STPG), by combining two principles from existing graph OOD methods. On one hand, we utilize the principle of environment diversity to handle the complex spatial relations in graph data, which is increasingly recognized as crucial in graph OOD research. On the other hand, we adopt the principle of invariance assumption to efficiently diversify only environment features while keeping causal ones intact. These two principles are embedded in our proposed diffIRM, which combines the merits of three graph OOD methods, namely, causality-based, invariant learning and data augmentation. First, we employ a GCN-based diffusion model to augment the input features. Second, we train a causal mask generator to identify causal features. Last, the augmented data, representing various environments, is used to define an invariance penalty that guides the training of the STPG model. The numerical experiment shows that diffIRM can identify the causal features. Using the SafeGraph, PeMS 04, and PeMS 08 datasets, we have found that diffIRM generates more accurate predictions of human mobility compared to baselines. Also, we demonstrate that diffIRM is model agnostic and can generate interpretable results.

There are two limitations of the proposed method. First, even though we use techniques like controlling the causal feature ratio to stabilize the training, the min-max problem remains more difficult to converge than other algorithms. This min-max problem arises from our data augmentation component. A future direction could be to explore a framework that does not involve data augmentation. Second, the current method does not consider the OOD issue within the test data, i.e., one test data segment may have a different distribution from other test segments. This could happen in cases where the data is highly temporally non-stationary. Tackling this problem is another future direction.

# Appendix A: Proofs

## A.1. Proof of Theorem 1

Using Assumption 1 and Condition 1, we have the following:

$$\mathcal{L}_{\text{cau}} <= \mathcal{L}^*(Y|X_{\text{cau}}, M^{(1:K)}) + \epsilon + \lambda \sum_{k=1}^{K} (\mathcal{L}^*(Y|X_{\text{cau}}, M) + \epsilon - \mathcal{L}^*(Y|X_{\text{cau}}, M^{(k)}) + \epsilon)$$

$$= (1 + 2\lambda K)\epsilon + \mathcal{L}^*(Y|X_{\text{cau}}, M^{(1:K)}) + \lambda \sum_{k=1}^{K} (\mathcal{L}^*(Y|X_{\text{cau}}, M^{(1:K)}) - \mathcal{L}^*(Y|X_{\text{cau}}, M^{(k)}))$$

$$= (1 + 2\lambda K)\epsilon + \mathcal{L}^*(Y|X_{\text{cau}})$$

$$\leq (1 + 2\lambda K)\epsilon + \mathcal{L}^*(Y),$$

which uses the property of Condition 1: as the generated mask does not impact causal features, $\mathcal{L}^*(Y|X_{\text{cau}}, M) = \mathcal{L}^*(Y|X_{\text{cau}})$. $\mathcal{L}^*(Y)$ is the optimal loss value with no predictor variables, e.g. variance of the squared error loss.

Similarly, we have the following using Assumption 1:

$$\mathcal{L}_{\text{env+}} >= \mathcal{L}^*(Y|X_{\text{env+}}, M^{(1:K)}) - \epsilon + \lambda \sum_{k=1}^{K} (\mathcal{L}^*(Y|X_{\text{env+}}, M^{(1:K)}) - \epsilon - \mathcal{L}^*(Y|X_{\text{env+}}, M^{(k)}) - \epsilon)$$

$$= -(1 + 2\lambda K)\epsilon + \mathcal{L}^*(Y|X_{\text{env+}}, M^{(1:K)}) + \lambda \sum_{k=1}^{K} (\mathcal{L}^*(Y|X_{\text{env+}}, M) - \mathcal{L}^*(Y|X_{\text{env+}}, M^{(k)}))$$

$$\geq -(1 + 2\lambda K)\epsilon + \lambda \sum_{k=1}^{K} (\mathcal{L}^*(Y|X_{\text{env+}}, M) - \mathcal{L}^*(Y|X_{\text{env+}}, M^{(k)})).$$

The third line uses the non-negative property of the loss function. Then, using Assumptions 2, 3, and 4, we have the following:

$$\mathcal{L}_{\text{env+}} >= \mathcal{L}^*(Y|X_{\text{env+}}, M) - \epsilon + \lambda \sum_{k=1}^{K} (\mathcal{L}^*(Y|X_{\text{env+}}, M) - \epsilon - \mathcal{L}^*(Y|X_{\text{env+}}, M^{(k)}) - \epsilon)$$

$$= -(1 + 2\lambda K)\epsilon + \mathcal{L}^*(Y|X_{\text{env+}}, M) + \lambda \sum_{k=1}^{K} (\mathcal{L}^*(Y|X_{\text{env+}}, M) - \mathcal{L}^*(Y|X_{\text{env+}}, M^{(k)}))$$

$$\geq -(1 + 2\lambda K)\epsilon + \lambda \sum_{k=1}^{K} (\mathcal{L}^*(Y|X_{\text{env+}}, M) - \mathcal{L}^*(Y|X_{\text{env+}}, M^{(k)})).$$

$$\geq -(1 + 2\lambda K)\epsilon + \lambda\delta \sum_{k=1}^{K} (\mathcal{L}^*(Y|X_s, M_s) - \mathcal{L}^*(Y|X_s, M_s^k))$$

$$\geq -(1 + 2\lambda K)\epsilon + \lambda\delta KC.$$

To let $\mathcal{L}_{\text{cau}} < \mathcal{L}_{\text{env+}}$, we can solve the following

$$(1 + 2\lambda K)\epsilon + \mathcal{L}^*(Y) \leq -(1 + 2\lambda K)\epsilon + \lambda\delta KC.$$

The solutions are $\lambda > \frac{\mathcal{L}^*(Y) + 2\epsilon}{\delta KC - 4K\epsilon}$ and $\epsilon < \frac{\delta C}{4K}$.

Next, for the case where only a fraction of invariant features $X_{\text{cau-}} \subset X_{\text{cau}}$ is included, we will prove $\mathcal{L}_{\text{cau}} < \mathcal{L}_{\text{cau-}}$ to ensure that our loss will guide the neural network to find complete causal features. In the previous step, we have shown that

$$\mathcal{L}_{\text{cau}} \leq (1+2\lambda)\epsilon + \mathcal{L}^*(Y|X_{\text{cau}}).$$

Similarly, using Condition 1 we have

$$\mathcal{L}_{\text{cau-}} \geq -(1+2\lambda K)\epsilon + \mathcal{L}^*(Y|X_{\text{cau-}}).$$

Then according to Assumption 2, we have

$$\mathcal{L}_{\text{cau-}} - \mathcal{L}_{\text{cau}} \geq -(2+4\lambda K) + \mathcal{L}^*(Y|X_{\text{cau-}}) - \mathcal{L}^*(Y|X_{\text{cau}})$$

$$\geq -(2+4\lambda K) + \gamma.$$

Thus, if $\epsilon < \frac{\gamma}{2+4\lambda K}$, we have

$$\mathcal{L}_{\text{cau-}} > \mathcal{L}_{\text{cau}}.$$

In conclusion, with $\lambda > \frac{\mathcal{L}(Y)+2\epsilon}{\delta C - 4K\epsilon}$ and $\epsilon < \min\{\frac{\delta C}{4K}, \frac{\gamma}{2+4\lambda K}\}$, we can get $\mathcal{L}_{\text{cau}} < \mathcal{L}'$, where $\mathcal{L}'$ is the value of loss function with regard to feature $X' \subset X$ with $X' \neq X_{\text{cau}}$. The proof is complete by noticing that $\epsilon$ can be chosen arbitrarily according to Assumption 1.

### A.2. Proof of Proposition

Now we have provided sufficient conditions to ensure the diffIRM can find the invariant features. We then continue to prove that Conditions 1 are also necessary.

Consider the following feature set

$$X_{\bar{v}} := \max_{|X'|}\{X' \subset X' : \mathcal{L}^*(Y|X', M) - \mathcal{L}^*(Y|X', M^{(k)}) = 0, \forall k \in \{1, \cdots, K\}\}.$$

Denote $\mathcal{L}_{\bar{v}}$ as the value of loss function associated with $X_{\bar{v}}$ By Assumption 1, for a given $\epsilon$ we can get

$$\mathcal{L}_{\bar{v}} \leq (1+2\lambda K)\epsilon + \mathcal{L}^*(Y|X_{\bar{v}}, M) + \lambda \sum_{k=1}^{K}(\mathcal{L}^*(Y|X_{\bar{v}}, M) - \mathcal{L}^*(Y|X_{\bar{v}}, M^{(k)}))$$

$$= (1+2\lambda K)\epsilon + \mathcal{L}^*(Y|X_{\bar{v}}, M)$$

$$\leq (1+2\lambda K)\epsilon + \mathcal{L}^*(Y).$$

On the other hand, suppose the augmentaion $M^{k'} \in \{M^k\}_{k=1}^{K}$ will impact causal features $X_{vk'}$ and all other augmentation does not impact causal features. Then, we have

$$\mathcal{L}_{\text{cau}} \geq -(1+2\lambda K)\epsilon + \mathcal{L}^*(Y|X_v, M) + \lambda \sum_{k=1}^{K}(\mathcal{L}^*(Y|X_v, M) - \mathcal{L}^*(Y|X_v, M^{(k)}))$$

$$\geq -(1+2\lambda K)\epsilon + \lambda \sum_{k=1}^{K}(\mathcal{L}^*(Y|X_v, M) - \mathcal{L}^*(Y|X_v, M^{(k)}))$$

$$\geq -(1+2\lambda K)\epsilon + \lambda\delta(\mathcal{L}^*(Y|X_{v^{k'}}, M) - \mathcal{L}^*(Y|X_{v^{k'}}, M^{k'})).$$

$$\geq -(1+2\lambda K)\epsilon + \lambda\delta C.$$

The proof is similar to that of Theorem 1. Thus, if we select $\epsilon < \delta C/4$ and $\lambda > \frac{\mathcal{L}^*(Y)+2\epsilon}{\delta C - 4K\epsilon}$, we have

$$\mathcal{L}_{\bar{v}} < \mathcal{L}_{\text{cau}}.$$

### A.3. Discussion on Condition 1

On one hand, given Assumption 2, the optimal strategy for the diffusion augmentor $G_\psi$ is to introduce the maximum amount of noise to the non-masked feature $X_{M^-} := X \odot (1 - M_{\text{cau}})$. This is because, by doing so, the information in $X_{M^-}$ is noised and thus will not help lower the objective function of the min-max game defined in Eq.. 4.1 according to Assumption 2. On the other hand, to lower the objective function, the causal mask generator is trained to find causal features $X_{\text{cau}}$. As the relation between $Y$ and causal features $X_{\text{cau}}$ is invariant across environments, including causal features can better lower the loss function. Thus the best strategy of the causal generator is to only exclude the environment features.

### Appendix B: More Experimental Details

**Regularization on Causal Feature Ratio.** Training the causal mask generator $T_\phi$ to identify causal features from scratch can be time-consuming. To address this, we implement a regularization on the ratio of causal features as follows:

$$\mathcal{L}_{\text{reg}}(\phi) = (\sum_{i=1}^{N}\sum_{j=1}^{\tau}\sum_{k=1}^{F}[M_{\text{cau}}]_{i,j,k}/N\tau F - \alpha)^2.$$

This regularization is incorporated into the loss function Eq. 4.5 to facilitate the training of $T_\phi$. In our experiments, we set $\alpha$ to 0.5, treating it as an initial estimate of causal features ratio. This approach accelerates the convergence of $T_\phi$, particularly in the initial stages, and becomes less influential as $T_\phi$ becomes well-trained. The primary purpose of this regularization is to prevent $T_\phi$ from converging prematurely to a local optimum where all features are incorrectly classified as either causal or environment features.

**Implementing Approximated diffIRM**. Implementing the penalty term in Eq. 4.6, which involves training a total of $K$ environment-specific predictors $f_{\theta_k}(\cdot)$, can be computationally intensive. To address this, we replace the penalty term with its first-order approximation as follows:

$$\hat{r}(\theta) = \sum_{k=1}^{K}\mathbb{E}_{(X,Y)\sim D}[\sum_{k=1}^{K}\nabla_\theta \ell_{\theta,\phi,\psi}(f_\theta(X \odot M_{\psi,\phi}^{(k)}), Y)/K].$$

For a more detailed explanation of this approximation, readers are referred to Arjovsky et al. (2019).

**Baseline Details.** The loss functions of baselines are as follows:

- Empirical Risk Minimization (ERM). ERM is a prevalent supervised learning approach that focuses on minimizing the average loss across training data without accounting for potential shifts in the environment. The objective function is given by

$$\mathcal{R}_{\text{ERM}}(\theta) = \mathbb{E}_{(X,Y)\sim \mathcal{D}}\ell(f_\theta(X), Y).$$

- IRM. Arjovsky et al. (2019) proposed this approach for learning causal and invariant features across different environments, which minimizes the worst-case loss over the environments under a linear constraint. The objective function is given by

$$\mathcal{R}_{\text{IRM}}(\theta) = \sum_{e=1}^{E} \mathbb{E}_{(X,Y)\sim \mathcal{D}_e}[\ell(f_\theta(X), Y)] + \lambda \sum_{e=1}^{E} \|\nabla_\theta \mathbb{E}_{(X,Y)\sim \mathcal{D}_e}[\ell(f_\theta(X), Y)]\|^2$$

where $\mathcal{D}_e$ is the distribution of the $e$-th environment, and $\lambda$ is a regularization coefficient.

- REx. For learning causal and invariant features across different environments, which minimizes the worst-case loss over the environments under a nonlinear constraint. The objective function is given by

$$\mathcal{R}_{\text{REx}}(\theta) = \min_{e=1,\ldots,E} \mathbb{E}_{(X,Y)\sim \mathcal{D}_e}[\ell(f_\theta(X), Y)] + \lambda \mathbb{E}_{(X,Y)\sim \mathcal{D}_e}[\ell(f_\theta(X), Y)]^2$$

where $\mathcal{D}_e$ is the data associated with the $e$-th environment, and $\lambda$ is a regularization coefficient. REx operates under the assumption that the environment is known, allowing $\mathcal{D}_e$ to be explicitly defined. In our implementation, we assume that each segment of historical data originates from a distinct environment.

- InvRat. Like REx, InvRat operates under the assumption that the environment is known. In our implementation, we similarly assume that each segment of historical data originates from distinct environments. The objective function is given by

$$\mathcal{R}_{\text{InvRat}}(\theta, \phi) = \sum_{e=1}^{E} \mathbb{E}_{(X,Y)\sim \mathcal{D}_e}[\ell(f_\theta(X), Y) + \alpha \ell(g_\eta(X), z_\theta(X))] + \beta \mathbb{E}_{(X,Y)\sim \mathcal{D}}[I(z_\theta(X), e|Y)]$$

where $\mathcal{D}_e$ is the data associated with the $e$-th environment, $\mathcal{D}$ is the joint distribution of all environments, $\theta$ and $\eta$ are the parameters of the prediction function $f$ and the rationalization function $g$, respectively, $z$ is the rationale extraction function, $\ell$ is a loss function, $I$ is the mutual information, and $\alpha$ and $\beta$ are loss function weights.